# Synthetic-to-Real Domain Bridging for Single-View 3D Reconstruction of Ships for Maritime Monitoring

Borja Carrillo-Perez[1], Felix Sattler[1], Angel Bueno Rodriguez[1], Maurice Stephan[1], and Sarah Barnes[1]

[1]German Aerospace Center (DLR), Institute for the Protection of Maritime Infrastructures, Germany

## ABSTRACT

Three-dimensional (3D) reconstruction of ships is an important part of maritime monitoring, allowing improved visualization, inspection, and decision-making in real-world monitoring environments. However, most state-of-the-art 3D reconstruction methods require multi-view supervision, annotated 3D ground truth, or are computationally intensive, making them impractical for real-time maritime deployment. In this work, we present an efficient pipeline for single-view 3D reconstruction of real ships by training entirely on synthetic data and requiring only a single view at inference. Our approach uses the Splatter Image network, which represents objects as sparse sets of 3D Gaussians for rapid and accurate reconstruction from single images. The model is first fine-tuned on synthetic ShapeNet vessels and further refined with a diverse custom dataset of 3D ships, bridging the domain gap between synthetic and real-world imagery. We integrate a state-of-the-art segmentation module based on YOLOv8 and custom preprocessing to ensure compatibility with the reconstruction network. Postprocessing steps include real-world scaling, centering, and orientation alignment, followed by georeferenced placement on an interactive web map using AIS metadata and homography-based mapping. Quantitative evaluation on synthetic validation data demonstrates strong reconstruction fidelity, while qualitative results on real maritime images from the ShipSG dataset confirm the potential for transfer to operational maritime settings. The final system provides interactive 3D inspection of real ships without requiring real-world 3D annotations. This pipeline provides an efficient, scalable solution for maritime monitoring and highlights a path toward real-time 3D ship visualization in practical applications. Interactive demo: https://dlr-mi.github.io/ship3d-demo/.

## 1. INTRODUCTION

Three-dimensional (3D) reconstruction of ships is increasingly important for maritime monitoring, offering enhanced situational awareness, improved safety, and operational insight for ports and authorities.[1,2] 3D models can allow more intuitive visualization, measurement, and analysis within interactive map environments, supporting a wide range of real-world maritime applications. For instance, 3D models can support maritime surveillance, vessel dimensions verification, or planning port logistics, particularly in scenarios requiring rapid interpretation of visual data.[1]

Traditional learning-based 3D reconstruction methods, especially those based on volumetric, mesh, or Neural Radiance Field (NeRF) representations,[3–5] are computationally expensive and require either multi-view supervision or slow inference, making them impractical for real-time maritime monitoring. Recently, Splatter Image[6] was introduced as an efficient alternative. By representing objects as sparse sets of 3D Gaussians,[7] rendered from single images, Splatter Image serves fast inference while maintaining reconstruction quality. This efficiency makes it well-suited for deployment in operational environments where speed and visual fidelity are needed.

Recent advances in single-view 3D reconstruction have achieved strong results for object-centric and indoor scenes, largely due to large-scale annotated datasets with multi-view 3D ground truth.[8,9] However, in maritime settings, there is a notable lack of real-world datasets containing 3D geometry or camera pose annotations for

Corresponding Author: borja.carrilloperez@dlr.de

ships. Existing maritime datasets, such as ShipSG,[10] provide high-quality images, ship segmentation masks, and metadata (e.g., geographic positions and ship length), but do not supply the 3D models needed for supervised learning of ship geometry. The absence of real-world 3D annotations remains a critical obstacle: most modern approaches cannot be directly applied to maritime monitoring. While domain adaptation and synthetic-to-real learning have shown promise for bridging the synthetic-to-real gap in other fields,[11,12] these techniques have seen limited application in maritime scenarios, where real-time inference and geospatial alignment are essential.

To address the limitations, we propose a pipeline for single-image 3D ship reconstruction, trained entirely on synthetic data, that can be used in real maritime monitoring settings. Synthetic datasets provide the 3D supervision needed for training, without relying on real-world annotations that are not available in maritime settings. Our approach fine-tunes the Splatter Image network for 3D ship reconstruction on existing and custom synthetic 3D ship datasets, then applies postprocessing and georeferencing to enable map-based 3D visualization of ships from real images, without requiring any real-world 3D ground truth. The pipeline supports interactive viewing and analysis of reconstructed ships on a web map, with placement and orientation based on scene metadata. We quantitatively evaluate reconstruction quality on synthetic data and show qualitative results on real ShipSG images, demonstrating the potential for transfer and practical use for maritime monitoring.

## 2. METHODS

### 2.1 Overview of the Pipeline

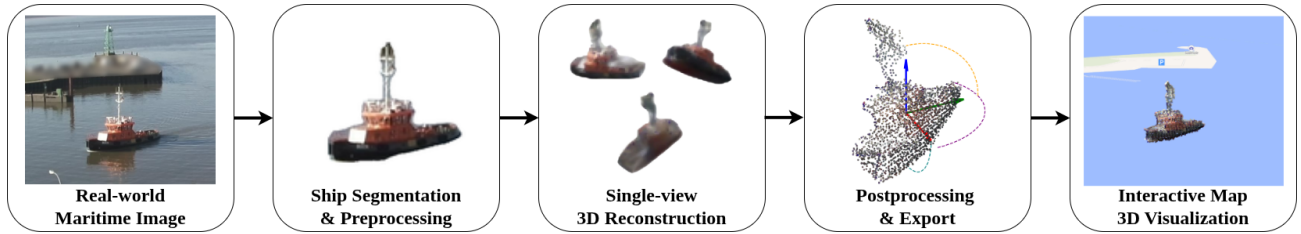

Figure 1. Overview of the proposed pipeline. Each block illustrates a step, from ship segmentation to 3D reconstruction and interactive map visualization.

Our proposed pipeline starts with an image captured by a maritime surveillance camera. The ship present is segmented from the background, and the image is preprocessed to ensure a standardized input format. This processed image is used for single-view 3D reconstruction trained on synthetic data. The resulting point cloud is exported and postprocessed to apply real-world scaling and orientation. The reconstructed 3D model is then georeferenced and visualized on an interactive map for user inspection. The following subsections describe each stage of the pipeline in more detail.

### 2.2 Real-world maritime images: ShipSG dataset

We use the ShipSG dataset[10] for training and evaluating the segmentation component of our pipeline. ShipSG is a publicly available dataset of real-world maritime surveillance imagery, collected from static cameras positioned at port infrastructure. The ground truth of each image in the ShipSG dataset contains annotated ship masks, ship class labels, and associated georeferencing metadata obtained from Automatic Identification System (AIS). AIS is an automated tracking system that uses transceivers on ships to continuously broadcast a vessel's identity, position (latitude and longitude), speed, and course, enabling other ships and maritime authorities to monitor and track vessel movements for navigation safety and traffic management.[13] The dataset includes this AIS-derived information, such as latitude, longitude, and ship length, for each annotated ship. The combination of instance segmentation, class, and georeferencing data is particularly suitable for our task, as it allows the presentation of detected and reconstructed ships on a georeferenced 3D map for maritime situational awareness. Furthermore, the dataset provides homography parameters for mapping ship masks to real-world coordinates, as well as Automatic Identification System (AIS) data, which is used to infer physical ship dimensions.

## 2.3 Ship Segmentation and Preprocessing

For ship detection and segmentation, we adopt an enhanced YOLOv8 architecture[14] that incorporates the scattering transform and Convolutional Block Attention Module (CBAM),[15] as described in.[16] The scattering transform provides translation-invariant, multi-scale feature extraction from the input image, while CBAM introduces channel and spatial attention mechanisms, improving detection and segmentation performance in challenging maritime environments. Further optimizations for embedded systems and inference speed were proposed in.[17] The ScatYOLOv8+CBAM model achieves state-of-the-art performance on ShipSG, reaching a mean Average Precision (mAP) of 75.39% with their fastest model,[17] and demonstrates robust segmentation of small ships.

We employ the segmentation masks generated by this model as input to the subsequent 3D reconstruction stage, due to the high accuracy and efficiency on real maritime data. After segmentation, each ship is extracted from the maritime scene and placed on a plain background to match synthetic training data conventions. The ship is centered and scaled to occupy 65% of the image area, ensuring consistent object scale across varying input conditions. The resulting standardized image is resized to $128 \times 128$ pixels before being input to the Splatter Image reconstruction network.

## 2.4 Single-view 3D Reconstruction with Splatter Image

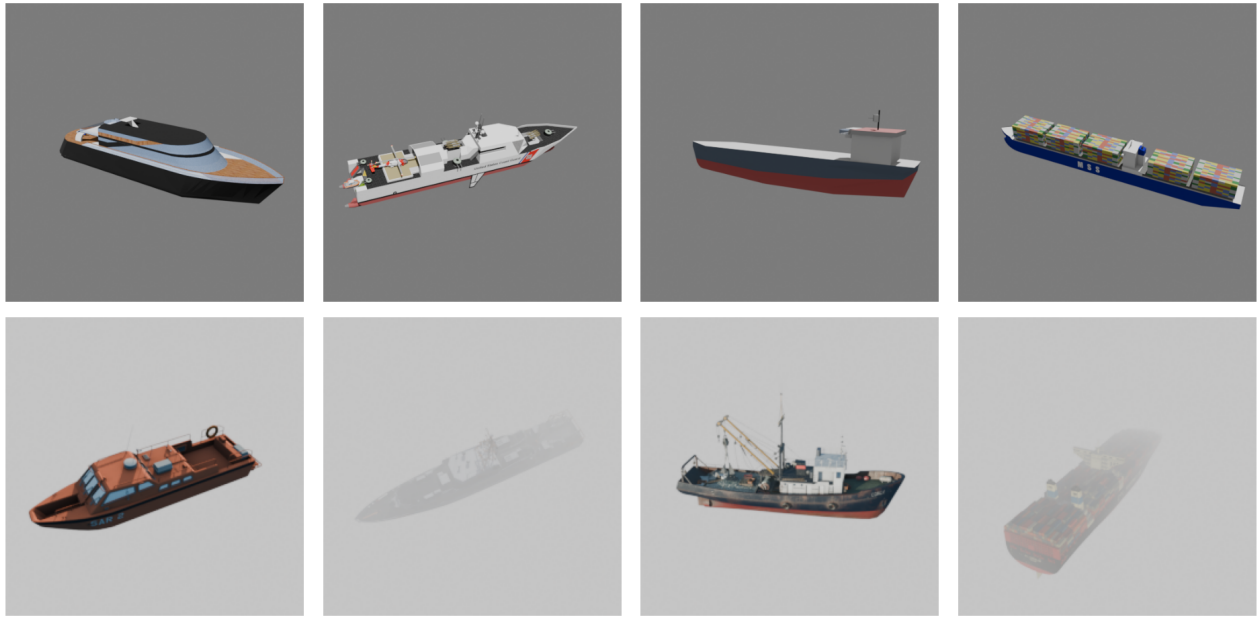

Figure 2. Qualitative comparison of randomly selected samples from the 'vessel' category of ShapeNet (top) and our proprietary dataset (bottom). Clearly noticeable is the improved realism in the bottom row with real-world lighting, environmental effects and physically-based materials.

For single-view 3D reconstruction, we employ the Splatter Image neural network.[6] Our approach begins with publicly available pretrained weights, originally trained on Objaverse, multi-category ShapeNet, CO3D (hydrants and teddybears), and ShapeNet (cars and chairs). To adapt the model to the maritime domain, we perform two-stage fine-tuning.

**First stage:** We fine-tune on the "vessel" category from ShapeNet, which adapts the learned features to ship-like geometries while preserving pretrained representations. For that, we use synthetic renders from multiple viewpoints and canonical orientations. We adapted the strategy proposed by Sitzmann et al.[18] and rendered each model from 16 viewpoints randomly sampled over a hemisphere that encloses a unit cube. All models are normalized, centered and aligned with the world axis for rendering. Input images are generated with $256 \times 256$ resolution to allow optional, higher-resolution training. In this work, input images were resized to $128 \times 128$ pixels and placed on a gray background to ensure consistency with real-data preprocessing. Figure 2, top row, shows four examples of randomly selected ShapeNet vessels. While texture and surface information might not be photorealistic, the general geometry and silhouette closely resemble the real-world data.

**Second stage:** We further fine-tune the model on a proprietary dataset of synthetic 3D ship models rendered in maritime conditions. Different commercially available 3D models were curated to include common vessel classes, like tugboats, container ships or tankers. We ensured that the materials of the 3D models were consistent under different lighting conditions by using a physically-based rendering-workflow.[19] Based on the ShapeNet workflow described above, the dataset was rendered using photorealistic rendering techniques in six real-world lighting setups. To further increase the realism of the images, four different versions with varying levels of fog were rendered to simulate the challenging weather conditions in the maritime domain. Figure 2, bottom row, shows four randomly selected samples from the proprietary dataset. Compared to the ShapeNet vessels it can clearly be seen how the detail in surface texture and lighting complexity has improved, leading to a more photorealistic appearance overall.

This two-stage fine-tuning process bridges the synthetic-to-real domain gap, using the single-view 3D reconstruction of ships from real maritime images without requiring real-world 3D ground truth. During inference, we use the Splatter Image network in its single-image reconstruction mode, consistent with our application scenario.

## 2.5 Postprocessing and Export

Following 3D reconstruction, the produced point cloud is postprocessed and exported for interactive visualization and downstream applications. The key steps are as follows:

**Real-world Scaling of 3D Models:** To provide physically meaningful models, the reconstructed point cloud is uniformly scaled so that its spatial extent along the z-axis (parallel to the water surface) matches the real-world ship length. Specifically, after canonical alignment, the extent along the z-axis is calculated as the difference between the maximum and minimum z-coordinates in the point cloud. A scale factor is then computed as the ratio of the target length (from AIS metadata in ShipSG) to the current z-extent, and all points in the cloud are multiplied by this factor. This assumes that all ships are consistently aligned with the z-axis following a fixed sequence of rotation and centering transformations applied during export. This procedure ensures that exported models are dimensionally consistent and geospatially meaningful for subsequent visualization.

**Point Cloud Orientation:** The reconstructed point cloud is recentered by subtracting the centroid (mean) from all point coordinates, ensuring predictable placement and alignment. We then apply a fixed sequence of rotation matrices so that the orientation of the exported 3D model consistently matches the viewpoint of the original input image. In other words, the 3D point cloud's orientation directly corresponds to the camera pose from which the image was taken. The interactive map viewer uses matching coordinate conventions, so the ship appears on the map as it did in the input image, including the real-world heading. By maintaining these conventions throughout the pipeline, the reconstructed ship's pose on the map accurately reflects its appearance in the source image.

**Opacity Threshold and Filtering:** To reduce visual noise, only points with an opacity value above a defined threshold are retained for export. This filters out low-confidence or spurious points, resulting in cleaner and more visually coherent reconstructions.

These steps collectively provide scale-accurate, consistently oriented, and visually reliable 3D models, facilitating downstream georeferenced visualization and analysis. The final postprocessed point cloud is exported in the standard Polygon File Format (PLY)[20] for compatibility with visualization tools and web viewers. The exported PLY files contain only the 3D coordinates $(x, y, z)$ and RGB color for each point, with the surface normals set to zero for compatibility. Additional per-point attributes used internally by Splatter Image, such as Gaussian scale (point size) and rotation, are omitted, as they are not required for standard point cloud visualization and would limit compatibility.

## 2.6 3D Ship Geospatial Visualization

The postprocessed 3D ship models are visualized within an interactive, web-based georeferenced map environment using a custom integration of Three.js and MapLibre GL JS.[21,22] The exported point cloud is rendered as a set of colored points within a 3D scene, embedded directly atop a basemap corresponding to real-world geography. This allows intuitive inspection, measurement, and situational awareness in a maritime context.

**Placement and Georeferencing:** Each ship is placed at geographic coordinates obtained from ShipSG/AIS metadata. To ensure accurate positioning, we use a homography-based mapping[10] to convert a key pixel from the ship mask (segmented by ScatYOLOv8+CBAM) to real-world latitude and longitude. The chosen pixel, as described in,[10] corresponds to the ship's AIS antenna location at the hull-water intersection. This ensures the 3D model is geospatially aligned with the location of the actual ship.

**Orientation Consistency:** The visualization process, as described in 2.5, ensures that each reconstructed ship retains the same orientation it had relative to the camera in the input image. As a result, ships are always displayed on the map in accordance with their appearance in the original image, and no attempt is made to recover or display their actual real-world heading.

This integrated geospatial visualization environment bridges the gap between deep learning-based 3D ship reconstruction and operational maritime situational awareness, supporting both human interpretation and downstream analytical workflows.

## 3. EXPERIMENTS AND RESULTS

### 3.1 Training Setup

We initialize the Splatter Image 3D reconstruction model with publicly available pretrained weights from the official Splatter Image repository.[6] All training and validation are performed entirely on synthetic datasets, as no 3D ground truth is available for real-world maritime images. We first fine-tune Splatter Image on the ShapeNet "vessel" category, using a 80% and 20% split of synthetic ship models for training and validation, respectively. A second fine-tuning stage is carried out on a proprietary synthetic dataset of rendered 3D ship models, also using a 80% and 20% split. See details in Table 1.

Table 1. Details of synthetic datasets used for training and validation.

| Dataset | # Train Models | # Val Models |
|---|---|---|
| ShapeNet (vessel) | 1549 | 310 |
| Custom Synthetic Ships | 268 | 68 |

Training uses the default Splatter Image hyperparameters, except for camera intrinsics set to $z_{near} = 0.01$, $z_{far} = 2.0$, and FOV = 45°. The training losses are the pixelwise L2 loss (i.e. mean squared error, MSE) and Learned Perceptual Image Patch Similarity loss (LPIPS[23]), with $\lambda_{\text{LPIPS}} = 0.01$.

### 3.2 Quantitative Results on Synthetic Data

We use the Structural Similarity Index (SSIM) and Peak Signal-to-Noise Ratio (PSNR) to evaluate the quality of the reconstructed images relative to ground truth. SSIM measures perceptual similarity and PSNR reflects overall reconstruction fidelity. Model selection is performed based on validation performance at $88k$ training iterations for the first fine-tuning, and $24k$ iterations for the second fine-tuning. The best SSIM achieved is 0.97, with PSNR of 39.37 dB. Table 2 summarizes the quantitative evaluation of our models on the synthetic validation sets, focusing on novel view synthesis. Both SSIM and PSNR are reported for images rendered from viewpoints not seen during training. Fine-tuning on the custom synthetic ship dataset yields improved performance compared to the initial ShapeNet (vessels) fine-tuning, confirming the benefit of additional domain-specific data for generalization to new viewpoints and ship geometries.

Table 2. Quantitative evaluation on synthetic validation data for novel view synthesis.

| Training Stage | SSIM | PSNR (dB) |
|---|---|---|
| ShapeNet (vessels) | 0.89 | 21.85 |
| Custom Synthetic Ships | 0.97 | 39.37 |

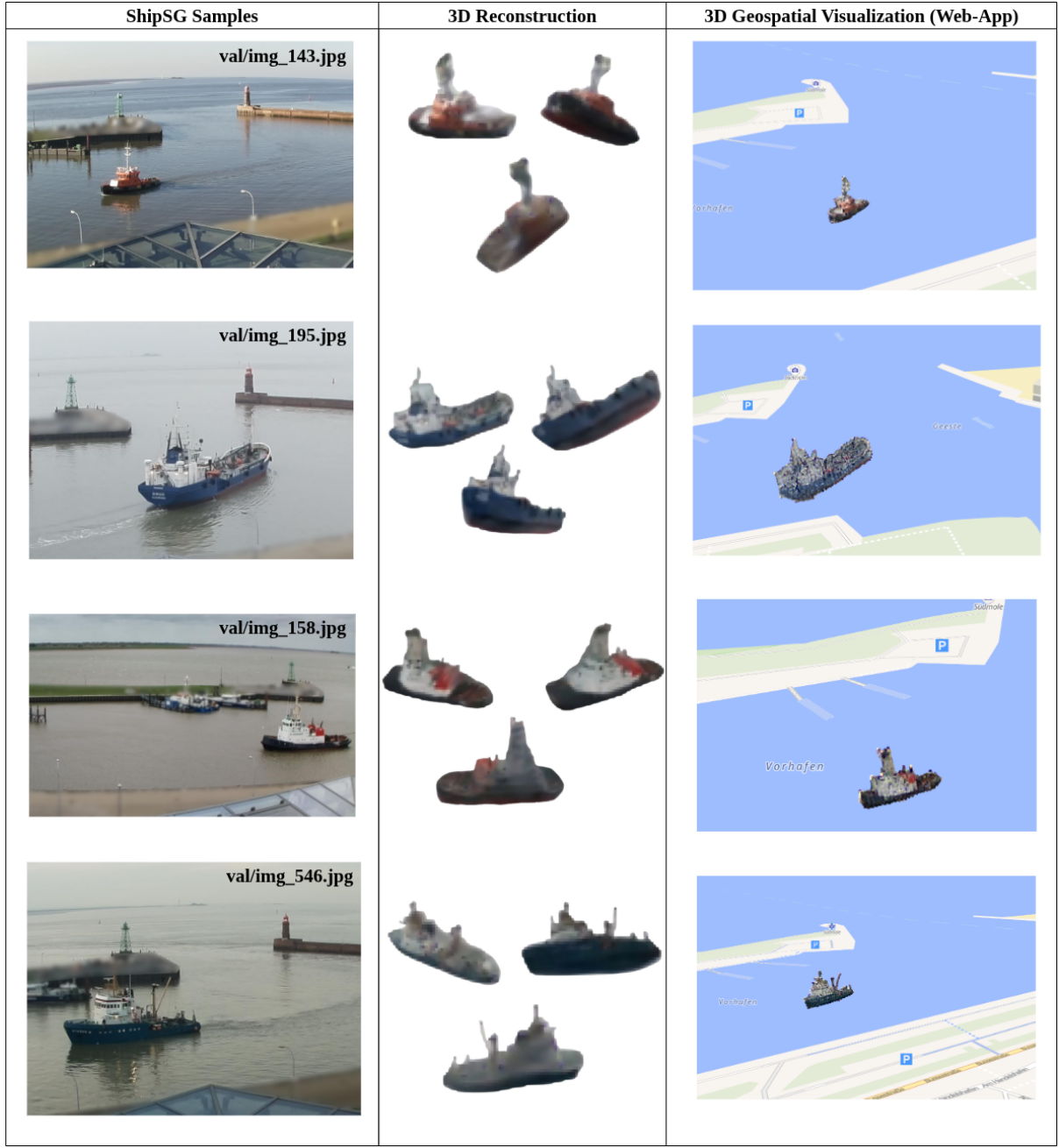

Figure 3. Qualitative results on real ShipSG data. Left: input images from ShipSG. Middle: 3D reconstruction using custom Splatter Image, trained on synthetic ships only. Right: view of interactive map visualization with correctly placed and oriented 3D models.

The results of our experiments shown in Table 2 are consistent and even present higher values compared with the metrics reported in the Splatter Image paper for their official pretrained model on their used datasets.[6] Their full model achieves SSIM 0.90 and PSNR of 22.25 dB. This confirms that our pipeline implementation faithfully reproduces baseline performance on standard benchmarks.

### 3.3 3D Geospatial Visualization on Real ShipSG Data

The trained pipeline is then applied to real ShipSG images to demonstrate its practical performance, as shown in Figure 1. As 3D ground truth is not available for real-world ships, results are presented qualitatively. Figure 3 shows examples of the full pipeline in operation, showing input images, reconstructed 3D models with Splatter Image, and geospatially accurate placement and orientation on the map.

As discussed in 2.6, during export, a fixed sequence of rotations is applied to the reconstructed point cloud, and these same rotation angles, $(\pi/2, \pi, \pi)$ for the X, Y, and Z axes, are used in the web-based viewer. These values were empirically determined to ensure visual consistency between the input camera view and the rendered ship orientation. We also empirically determined the opacity threshold, which we set to $-3.0$, to suppress low-confidence points while preserving reconstruction fidelity. Together, these steps ensure that each ship appears on the interactive map as it did in the input image, maintaining visual and spatial consistency throughout the pipeline, as described in Section 2.6.

Users can interact with the map-based viewer to inspect each ship from multiple angles and perform direct measurements using the integrated tool. Placement is determined using AIS-derived coordinates and homography-based mapping as described in the methods (see Section 2.5). The 3D visualization preserves the ship's appearance and viewpoint from the input image.

This demonstrates the end-to-end applicability of the proposed method for maritime situational awareness, bridging synthetic training and real-world deployment without requiring real 3D ground truth.

## 4. CONCLUSION AND FUTURE WORK

In this paper, we present a pipeline for single-view 3D reconstruction of ships, trained entirely on synthetic data. Our system can reconstruct ships from real maritime images and display the results as georeferenced 3D models on an interactive web map. By using domain-adapted deep learning and postprocessing, our method bridges the gap between synthetic and real data, overcoming the lack of real-world 3D ground truth. The system supports practical deployment in maritime situational awareness, since the models achieve good reconstruction accuracy on synthetic validation data and give visually realistic results on real ShipSG images, confirming the potential for transfer to practical use. Some challenges remain with highly complex or occluded ships, but the method provides a straightforward path toward scalable, operational 3D monitoring without needing real 3D annotations.

While we demonstrate plausible reconstructions on real maritime images, quantitative performance on real data remains unvalidated due to the absence of real 3D ground truth. Addressing this limitation is a key direction for future work. For example, generating real-world 3D ground truth with LiDAR and multi-camera systems would allow more comprehensive validation and possibly enable supervised domain adaptation to real maritime imagery. Additional future work includes integrating 3D ship reconstructions with higher-fidelity photogrammetric or LiDAR-based digital twins for improved map detail. We also plan to experiment with additional data sources and image modalities to boost robustness in diverse conditions.